\pdfoutput=1

\documentclass[11pt]{article}

\usepackage{acl}

\usepackage{times}
\usepackage{latexsym}
\usepackage{bm}
\usepackage{color,colortbl}
\usepackage{stfloats}
\usepackage{subfigure}
\usepackage{enumitem}
\usepackage{booktabs}
\usepackage{nccmath}
\usepackage{multirow}
\usepackage{subfigure}
\usepackage{makecell}
\usepackage{todonotes}
\usepackage{latexsym}
\usepackage{graphicx}
\usepackage{float}

\newcommand\xdownarrow[1][2ex]{%
   \mathrel{\rotatebox{90}{$\xleftarrow{\rule{#1}{0pt}}$}}
}

\usepackage[T1]{fontenc}

\usepackage[utf8]{inputenc}

\usepackage{microtype}

\usepackage{soul}

\title{Human Judgement as a Compass to Navigate Automatic Metrics for Formality Transfer}

\author{
Huiyuan Lai, Jiali Mao, Antonio Toral, Malvina Nissim\\
CLCG, University of Groningen / The Netherlands\\
\texttt{\{h.lai, jiali.mao, a.toral.ruiz, m.nissim\}@rug.nl}
}

\begin{document}
\maketitle
\begin{abstract}
Although text style transfer has witnessed rapid development in recent years, there is as yet no established standard for evaluation, which is performed using several automatic metrics, lacking the possibility of always resorting to human judgement.
We focus on the task of formality transfer, and on the three aspects that are usually evaluated: style strength, content preservation, and fluency. To cast light on how such aspects are assessed by common and new metrics, we run a human-based evaluation and perform a rich correlation analysis. 
We are then able to offer some recommendations on the use of such metrics in formality transfer, also with an eye to their generalisability (or not) to 
related tasks.\footnote{Our analysis code, literature list for Figure~1, and all data are available at \url{https://github.com/laihuiyuan/eval-formality-transfer}.}

\end{abstract}

\section{Introduction}
Text style transfer (TST) is the task of automatically changing the style of a given text while preserving its style-independent content, or theme. Quite different tasks, and thus quite different types of transformations, traditionally fall under the TST label. For example, given the sentence ``\emph{i like this screen, it's just the right size...}'', we may produce its negative counterpart ``\emph{i hate this screen, it is not the right size}'' for the task defined as \textit{polarity swap}~\citep{Shen-2017, li-etal-2018}, or turn it into the formal ``\emph{I like this screen, it is just the right size.}'' for the task called \textit{formality transfer}~\citep{rao-tetreault-2018}.

\begin{figure}[t]
    \centering
    \includegraphics[scale=0.48]{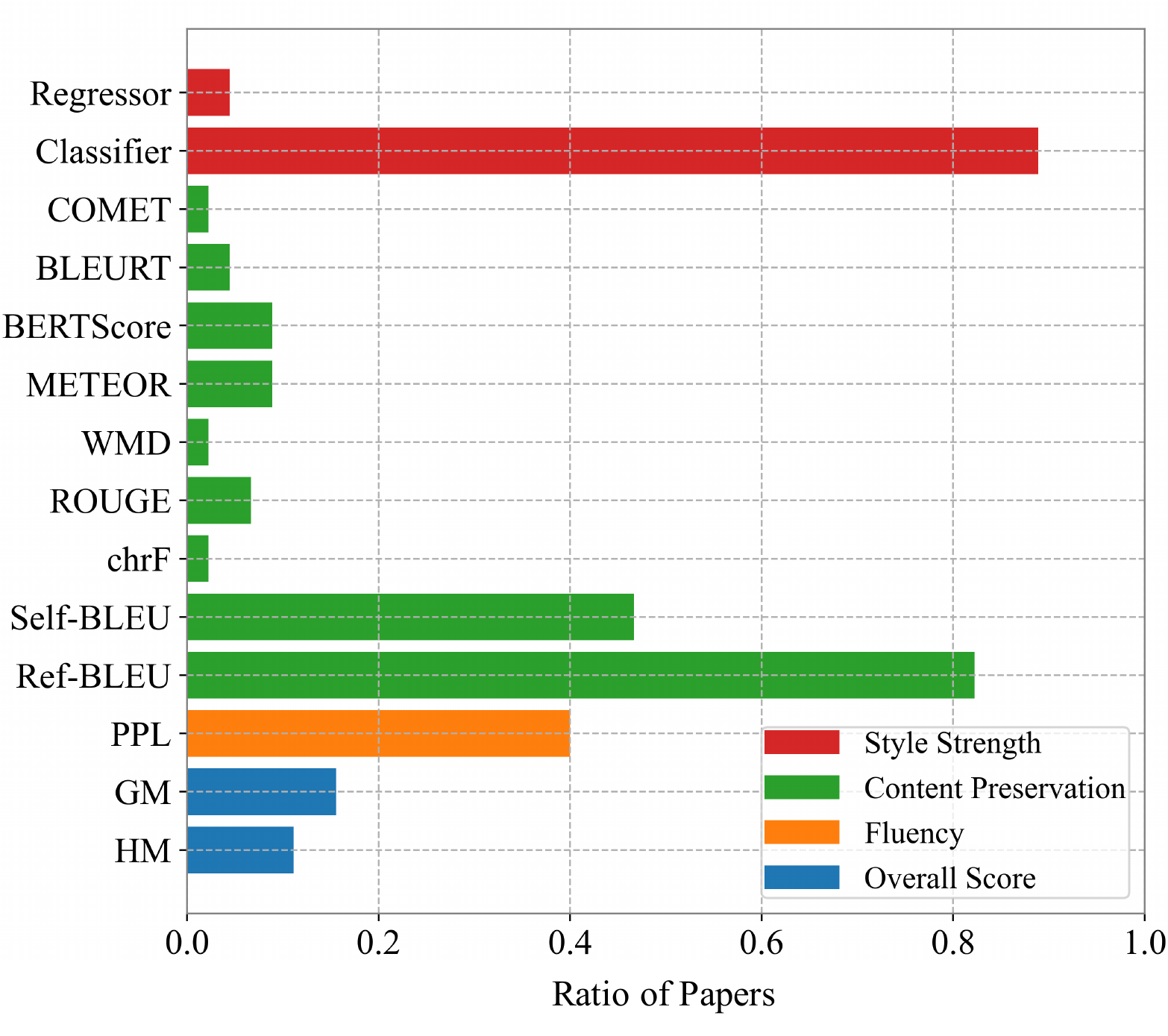}
    \caption{Automatic evaluation metrics in 45 ACL Anthology papers focusing on style transfer and its evaluation in terms of (i) style strength: regressor and classifier; (ii) content preservation: COMET, BLEURT, BERTScore, METEOR, WMD, ROUGE, chrF, Self-BLEU (source-based BLEU) and Ref-BLEU (reference-based BLEU); (iii) fluency: PPL (perplexity); and (iv) overall score:  HM (harmonic mean) and GM (geometric mean).}
    \label{fig:metrics}
\end{figure}

For the transfer to be considered successful, the output must be written (i) in the appropriate target style; (ii) in a way such that the original content, or theme, is preserved; and (iii) in proper language, hence fluent and grammatical (relative to the desired style). These aspects to be evaluated are usually defined as (i) \textit{style strength}, (ii) \textit{content preservation}, and (iii) \textit{fluency}, and automatic evaluation metrics are used accordingly, lacking the possibility of using human judgement for any given experiment. Figure~\ref{fig:metrics} shows a survey of such metrics (organised by aspect) as used in 45 papers published over the last three years in the ACL Anthology, which focus on TST in general. A classifier or a regressor is used to assess style strength, a variety of content-based metrics target content preservation,  perplexity is used to measure fluency, and some overall metrics combining content and style are often reported. 


In spite of the attempts to perform careful automatic evaluation, and of some works studying specific aspects of it, such as traditional metrics for polarity swap~\citep{tikhonov-etal-2019-style, mir-etal-2019-evaluating}, content preservation for formality transfer~\citep{yamshchikov2020styletransfer}, and a recent attempt at correlating automatic metrics and human judgment for some aspects of multilingual formality transfer~\citep{briakou-etal-2021-evaluating},  
the community has not yet reached fully shared standards in evaluation practices. 
We believe this is due to a concurrence of factors. 

First, different tasks are conflated under the TST label while they are not exactly the same, and evaluation is a serious issue. 
\citet{lai-etal-2021-generic} have shown that polarity swap and formality transfer cannot be considered alike especially in terms of content preservation, as in the former the meaning of the output is expected to be the opposite of the input rather than approximately the same. Hence, it is difficult to imagine that the same metric would capture well the content aspect in both tasks. 

Second, the evaluation setting is not necessarily straightforward: if the content of the input has to be preserved in the output, the quality of the generated text can be assessed either against the input itself or against a human-produced reference, specifically crafted for evaluation. However, not all metrics are equally suitable for both assessments. For instance, BLEU~\citep{papineni-etal-2002-bleu} is the metric most commonly used for evaluating content preservation (Fig.~\ref{fig:metrics}). Intuitively, this $n$-gram based metric should be appropriate for comparing the output and the human reference, but is much less suitable for comparing the model output and the source sentence, since the whole task is indeed concerned with changing the surface realisation towards a more appropriate target style. 
On the contrary, neural network-based metrics should also work between the model output and the source sentence. This leads to asking what the best way is to use and possibly combine these metrics under which settings. Closely related to this point, it is not fully clear what the used metrics actually measure and what desirable scores are. For example, comparing source and reference for metrics that measure content similarity should yield high scores, but we will see in our experiments that this is not the case.
Recent research has only compared using the reference and the source sentence for one metric: BLEU~\citep{briakou-etal-2021-evaluating}, and introduced some embeddings-based metrics only to compare the output to the source. A comprehensive picture of a large set of metrics in the two different evaluation conditions (output to source and output to reference) is still missing and provided in this contribution.

Lastly, and related to the previous point, it is yet unclear whether and how the used metrics correlate to human judgements under different conditions (e.g.\ not only the given source/reference used for evaluation but also different transfer directions, as previous work has assessed human judgement over the informal to formal direction \citep{briakou-etal-2021-evaluating} only), and how they differ from one another. This does not only affect content preservation, as discussed above, but also style strength and fluency.

Focusing on formality transfer, where the aspect of content preservation is clear, we specifically pose %
the following research questions:

\begin{itemize}[leftmargin=*]
\itemsep 0in
\item \textbf{RQ1} What is the difference in using a classifier or a regressor to assess style strength and how do they correlate with human judgement?

\item \textbf{RQ2} How do different content preservation metrics fare in comparison to human judgement, and how do they behave when used to compare TST outputs to source or reference sentences?

\item \textbf{RQ3} Is fluency well captured by perplexity, and what if the target style is informal?

\end{itemize}

\noindent To address these questions 
we conduct a human evaluation for a set of system outputs, collecting judgments over the three evaluation aspects, and unpack each of them by means a thorough correlation analysis with automatic metrics. 

\paragraph{Contributions} Focusing on formality transfer, we offer a comprehensive analysis of this task and the nature of each aspect of its evaluation. 
Thanks to the analysis of correlations with human judgements, we uncover which automatic metrics are more reliable for evaluating TST systems and which metrics might not be suitable for this task under specific conditions.
Since it is not feasible to always have access to human evaluation, having a clearer picture of which metrics better correlate with human evaluation is an important step towards a better systematisation of the task's evaluation.

\section{Related Work}
\label{sec:related-work}
\paragraph{Text Style Transfer} In the recent tradition of TST, many related tasks have been proposed by researchers.~\citet{xu-etal-2012-paraphrasing} employ machine translation techniques to transform modern English into Shakespearean English.
~\citet{sennrich-etal-2016-controlling} propose a task that aims to control the level of politeness via side constraints at test time. Polarity swap~\citep{Shen-2017, li-etal-2018-delete} is a task of transforming sentences, swapping their polarity while preserving their theme. Political slant is the task that preserves the intent of the commenter but modifies their observable political affiliation~\citep{prabhumoye-etal-2018-style}. Formality transfer is the task of reformulating an informal sentence into formal (or viceversa)~\citep{rao-tetreault-2018, briakou-etal-2021-ola}. ~\citet{cao-etal-2020-expertise} propose an expertise style transfer that aims to simplify the professional language in medicine to the level of laypeople descriptions using simple words.~\citet{jin2021deep} provide an overview for different TST tasks.

\paragraph{Automatic Evaluation} 
In Figure~\ref{fig:metrics} we see that more than 80\% of papers employ a style classifier to assess the attributes of transferred text for the aspect of style strength. 
For content preservation,
BLEU is by far the most popular automatic metric, but recent work has also employed other metrics, including string-based (e.g.\ METEOR~\citep{mir-etal-2019-evaluating, lyu-etal-2021-styleptb, briakou-etal-2021-evaluating}) and neural-based (e.g.\ BERTScore~\citep{reid-zhong-2021-lewis, lee-etal-2021-enhancing, briakou-etal-2021-evaluating}). 
In order to further increase the capturing of semantic information beyond the lexical level, 
\citet{lai-etal-2021, lai-etal-2021-generic} recently also employed BLEURT~\citep{sellam-etal-2020-bleurt} and COMET~\citep{rei-etal-2020-comet} to evaluate their systems. These \textit{learnable metrics} attempt to directly optimize the correlation with human judgments, and have shown promising results in  machine translation evaluation. For fluency, a language model (LM) trained on the training data is used to calculate the perplexity of the transferred text \citep{john-etal-2019-disentangled, sudhakar-etal-2019-transforming, huang-etal-2020-cycle}. 
Geometric mean and harmonic mean of style accuracy and BLEU are often used for overall performance~\citep{xu-etal-2018-unpaired, fuli-2019, krishna-etal-2020-reformulating, lai-etal-2021-generic, lai-etal-2021}.

\paragraph{Evaluation Practices} 
Although some previous work has run correlations of human judgements and automatic metrics~\citep{rao-tetreault-2018, fuli-2019}, this was not the focus of the contribution and no deeper analysis or comparison was run. On the other hand,~\citet{yamshchikov2020styletransfer} examined 13 content-related metrics in the context of formality transfer and paraphrasing, and show that none of the metrics is close enough to the human judgment. 
~\citet{briakou-etal-2021-evaluating} have recently evaluated automatic metrics on the task of multilingual formality transfer against human judgement.
We also examine automatic metrics in terms of correlation with human judgement, but there are some core differences between our contribution and their work. 
First, for style strength, they focus on comparing two different architectures in a cross-lingual setting using the correlation on human judgement for regression, and they do not provide this analysis for style classification, rather an evaluation against the gold label. In contrast, we adopt an architecture that provides regression and classification comparisons in fitting human judgments. Second, regarding content, \citet{briakou-etal-2021-evaluating} focus on similarity (and therefore metrics) to the source sentence, while we stress the importance of triangulation also with the reference\footnote{Although the reference is not always available, using it in studying evaluation metrics in comparison with how they behave when the source is used provides insights into the overall behaviour of such metrics and how they should best be employed even in the absence of a reference.}. Also, we introduce two learnable metrics in the evaluation setup, which correlation with human judgement shows to be the most informative. Third, they compare perplexity, likelihood, and pseudo-likelihood scores for fluency evaluation, while we provide a deeper evaluation of just perplexity considering though the two directions (\citet{briakou-etal-2021-evaluating} evaluate only the informal-to-formal direction) and highlight differences that point to a potential benefit in using different approaches or evaluation strategies for the two directions.

In addition, we (i) use a continuous scale setting for human judgement which, unlike a discrete Likert scale, allows to normalize judgments~\citep{graham-etal-2013-continuous}, hence increasing homogeneity of the assessments; 
(ii) evaluate eight existing, published systems of different sorts (including state-of-the-art models) for both transfer directions, thereby potentially enabling a reconsideration of results as reported in previous work; 
(iii) study the nature of each evaluation aspect and the corresponding automatic metrics, analyzing the differences in the correlation between metric and human judgements that might arise under different conditions (e.g.\ looking at high-quality systems).

\begin{figure}[t]
    \centering
    \includegraphics[scale=0.41]{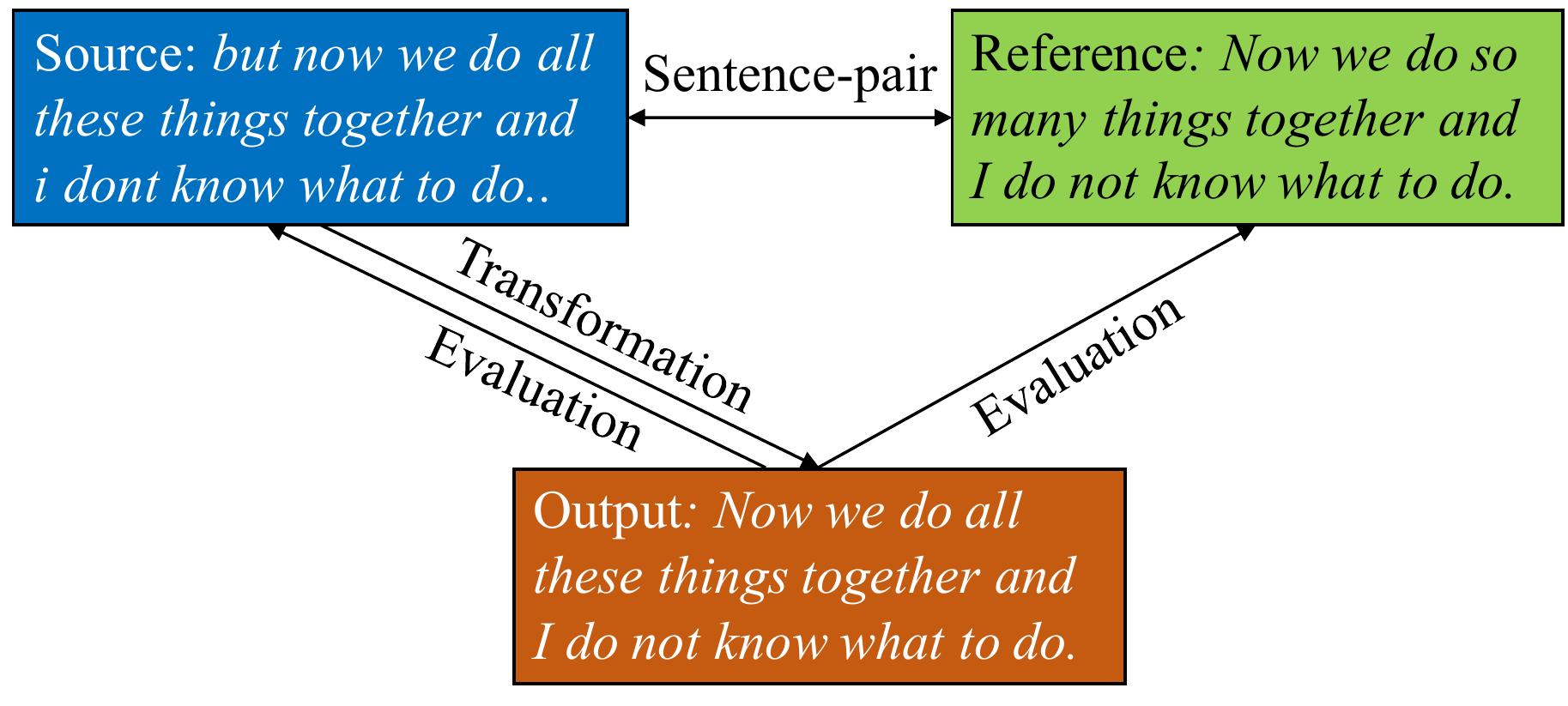}
    \caption{Alignment, transformation, evaluation pairs.}
    \label{fig:illustration}
\end{figure}

\section{Data}
\label{sec:data}
We use GYAFC~\cite{rao-tetreault-2018}, a formality transfer dataset for English that contains aligned formal and informal sentences from two domains: Entertainment \& Music and Family \& Relationships. 
Figure~\ref{fig:illustration} shows an example for alignment, transformation, and evaluation relations between input, output, and reference.
We run a human evaluation and a battery of automatic metrics on a selection of human- and machine-produced texts. 

\paragraph{Source and Reference Texts}
The source and reference texts we use are from the Family \& Relationships domain. The test set contains 1,332 and 1,019 sentences in ``informal to formal'' and ``formal to informal'' directions, respectively. There are four human references for each test sentence. We randomly select 80 source sentences (40 for each transfer direction) from the test set, as well as their corresponding human references. For each source sentence, we obtain the corresponding transformations as produced by eight different systems. 

\paragraph{System Outputs}
The evaluation results are often affected by the system's outputs, since if the evaluated systems are of different types, they may exhibit different error patterns so that various automatic evaluation metrics can be differently sensitive to these patterns~\citep{ma-etal-2019-results, mathur-etal-2020-results}. To fully examine the evaluation methods, the systems we use are all from previous work, both supervised and unsupervised approaches.\footnote{Details of the systems are in Appendix~\ref{app:formality-system}.} Overall, the eight systems yield a total of 640 output sentences (80 per system, 40 in each direction).

\section{Methodologies}

\subsection{Human Evaluation}
\label{sub:human}


To facilitate the annotation and obtain a manageable size for each annotator, we split the 80 source sentences (Section~\ref{sec:data}) into four different surveys with 20 sentences each (10 for each transfer direction), and their corresponding system outputs plus one reference. 

We recruited eight 
highly proficient English speakers for this task, i.e. two per survey, so that 
two annotations for each target sentence can be collected; from these we can use the average score assigned, and also calculate inter-annotator agreement.
The task is to rate the transferred sentence on a continuous scale (0-100), inspired by Direct Assessment~\cite{graham-etal-2013-continuous, Graham2015CanMT}, in terms of three evaluation aspects: (i) style strength (does the transformed sentence fit the target style?); (ii) content preservation (is the content of the transformed sentence the same as the original sentence?), and (iii) fluency (considering the target style, could the transformed sentence have been written by a native speaker?). 


Before starting the rating task, we provided annotators with detailed guidelines and examples of transformed sentences along with plausible assessments for each aspect.\footnote{Screenshots of our annotation guidelines and interface are in  Appendix~\ref{app:interface}.} We also reminded the annotators that such examples are only indicative of what we believe to be plausible judgements but there are many possible correct answers, of course.

\subsection{Automatic Evaluation}
\label{sub: metric}
We test a wide range of commonly used as well as new automatic metrics on the three  aspects.

\paragraph{Style Strength} The most commonly used method for assessing style strength is a style classifier, with the problem cast as a binary classification task (formal vs informal in formality transfer). \citet{briakou-etal-2021-evaluating} have recently shown that a style regressor fine-tuned with English rating data correlates better with human judgments in other languages (Italian, French, and Portuguese). To run a proper comparison, we use BERT~\citep{devlin-etal-2019-bert} as our base model, and fine-tune it with style labelled data (GYAFC) and the rating data of PT16~\citep{pavlick-tetreault-2016-empirical} to obtain a style classifier (C-GYAFC) and a regressor (R-P16), respectively. Following~\citet{rao-tetreault-2018}, we collect sentences from PT16 with human rating from -3 to +1 as informal and the rest as formal, and train a style classifier on them (C-PT16). C-GYAFC and C-PT16 achieve an accuracy of 94.4\% and 58.6\% on the test sets, respectively.

\paragraph{Content Preservation} 
We consider the following metrics, including both surface-based and embedding-based approaches: \footnote{The implementation details for automatic metrics are in Appendix~\ref{app:details-metrics}.}

\begin{itemize}[leftmargin=*]
\itemsep 0in
\item BLEU~\citep{papineni-etal-2002-bleu} It compares a given text to others (reference) by using a precision-oriented approach based on $n$-gram overlap;

\item chrF~\citep{popovic-2015-chrf} It measures the similarity of sentences using the character $n$-gram F-score;

\item ROUGE~\citep{lin-2004-rouge} It compares a given text to others (human reference) by using $n$-gram/the longest co-occurring in sequence overlap and a recall-oriented approach;

\item WMD~\citep{pmlr-v37-kusnerb15} It measures the dissimilarity between two texts as an optimal transport problem which is based on word embedding.

\item METEOR~\citep{banerjee-lavie-2005-meteor} It computes the similarity score of two texts by using a combination of unigram-precision, unigram-recall, and some additional measures like stemming and synonymy matching.

\item BERTScore~\citep{bert-score} It computes a similarity score for each token in the candidate sentence with each token in the reference sentence. Instead of exact matches, it computes token similarity using contextual embeddings. 

\item BLEURT~\citep{sellam-etal-2020-bleurt} It is a learned evaluation metric based on BERT \citep{devlin-etal-2019-bert}, trained on human judgements. It is trained with a pre-training scheme that uses millions of synthetic examples to help the model generalize.

\item COMET~\citep{rei-etal-2020-comet}  It is a learnable metric which leverages cross-lingual pretrained language modeling resulting in multilingual machine translation evaluation models that exploit both source and reference sentences.
\end{itemize}

\noindent For assessing content preservation in the output, we can exploit both the source and the reference (see Fig.~\ref{fig:illustration}). 
When comparing our output to the source, we want to answer the following question: (a) how close in content is the generated text to the original text?, which addresses naturally the content preservation aspect of the task. When comparing our output to the human-produced reference, we want to answer a different question: (b) how similar is the automatically generated text to the human written one? Both are valid strategies, but by answering  different questions they are likely to react differently to, and require, different metrics.

The advantages of the (a) approach are that evaluation is possible even without a human reference, it is the most natural way of assessing the task, and it does not incur reference bias~\citep{fomicheva-specia-2016-reference}. 
The core problem lies in the use and interpretation of metrics: surface-based metrics (like BLEU) would score highest if nothing has changed from input to output (if the model doesn't perform the task, basically), so aiming for a high score is pointless. A very low score is undesirable, too, however. For more sophisticated metrics, the problem is similar in the sense the highest score would be achieved if the two texts are identical, but since it is not fully clear what they measure exactly in terms of similarity, what to aim for isn't straightforward (an indication is provided by using metrics to compare source and reference).

The main advantage of the (b) approach is that metrics can be used in a more standard way: tending to the highest possible score is good for any of them, since getting close to the human solution is desirable. However, the gold reference is only one of many possible realisations, and while high scores are good, low scores can be somewhat meaningless, as proper meaning-preserving outputs may be very different from the human-produced ones, especially at surface level.

While we have as yet no specific solution to this, this study contributes substantially to a better understanding of automatic metrics, especially for content preservation, possibly leading to a combined metric which considers mainly the source, and possibly the reference(s) in a learning phase.

\paragraph{Fluency} In formality transfer, both informal and formal outputs must be evaluated. Intuitively, the latter should be more fluent and grammatical than the former so that evaluating the fluency of informal sentences might be more challenging, both for humans and automatic metrics. We use the perplexity of the language model GPT-2~\citep{radford-2019} fine-tuned with style labelled texts. Specifically, we fine-tune two GPT-2 models on informal sentences and formal sentences respectively, and then we use the target-style model to calculate the perplexity of the generated sentence. Finally, we provide a separate correlation analysis between automatic metrics and human judgements for the two transfer directions.

\subsection{Correlation Methods}

\paragraph{Pearson Correlation}
We employ Pearson correlation ($r$) as our main evaluation measure for system-/segment-level metrics:
\begin{equation}
\label{eq:pearson}
r = \frac{\sum_{i=1}^{n} (H_{i} - \bar{H}) (M - \bar{M})}
    {\sqrt{\sum_{i=1}^{n} (H_{i} - \bar{H})^2 \sum_{i=1}^{n} (M_{i} - \bar{M})^2}}
\end{equation}
where $H_{i}$ is the human assessment score, $M_{i}$ is the corresponding score as predicted by a given metric. $\bar{H}$ and $\bar{M}$ are the their means, respectively.

\paragraph{Kendall’s Tau-like formulation}
We follow the WMT17 Metrics Shared Task \citep{bojar-etal-2017-results} and take the official Kendall’s Tau-like formulation, $\tau$, as the our main evaluation measure for segment-level metrics. A true pairwise comparison is likely to lead to more stable results for segment-level  evaluation \citep{vazquez-Alvarez-etal}. The Kendall’s Tau-like formulation $\tau$ is as follows:

\begin{equation}
\label{eq:tau}
\tau = \frac{Concordant - Discordant}
    {Concordant + Discordant}
\end{equation}
Where $Concordant$ is the number of times for which a given metric suggests a higher score to the ``better'' hypothesis judged by human and $Discordant$ is the number of times for which a given metric suggests a higher score to the ``worse'' hypothesis judged by human.

Most automatic metrics, like BLEU, aim to achieve a strong positive correlation with human assessment, with the exception of WMD and perplexity, where the smaller is better. 
We thereby employ absolute 
correlation value for WMD and perplexity in the following analysis. 

\begin{table}[t]
\setlength{\tabcolsep}{5pt}
\footnotesize
\resizebox{\linewidth}{!}{%
\centering
\begin{tabular}{p{1.2cm}ccccc}
\toprule
 \textbf{Survey} & \textbf{N} & \textbf{Content} & \textbf{Style} & \textbf{Fluency} & \textbf{Overall}\\
 \hline
 Survey 1 & 160 & 0.90 & 0.45 & 0.71 & 0.70\\
 Survey 2 & 160 & 0.84 & 0.48 & 0.63 & 0.66\\
 Survey 3 & 160 & 0.83 & 0.68 & 0.70 & 0.72\\
 Survey 4 & 160 & 0.81 & 0.62 & 0.63 & 0.68\\
 \hline
 Overall  & 640 & 0.86 & 0.52 & 0.66 & 0.70\\
\bottomrule
\end{tabular}}
\caption{\label{tab:iaa} 
Inter-Annotator Agreement.}
\end{table}

\section{Results and Analysis}
In this section, we first measure the inter-annotator agreement of the human evaluation, then discuss both system-level and sentence-level evaluation results on the three aforementioned evaluation aspects, so as to provide a different perspective on the correlation between automatic metrics and human judgements under different conditions.

\subsection{Inter-Annotator Agreement}

There are two human judgements for each sentence and we measure their inter-annotator agreement (IAA) by computing the Pearson Correlation coefficient, instead of the commonly used Cohen's $K$, since judgements are given on a continuous scale. 

Table~\ref{tab:iaa} presents the results of IAA for each aspect in each single survey and overall. Across the four surveys annotators have the highest agreement on the content aspect, followed by fluency, with style yielding the lowest scores, suggesting that annotators have more varied perceptions of sentence style than content. 
Overall, we achieve reasonable 
agreement for all surveys and evaluation aspects.

\begin{table}[t]
\setlength{\tabcolsep}{5pt}
\resizebox{\linewidth}{!}{%
\footnotesize
\centering
\begin{tabular}{p{2.1cm}cccc}
\toprule
  & \textbf{N} & \textbf{R-PT16} & \textbf{C-PT16} & \textbf{C-GYAFC} \\
 \hline
  \multirow{1}{*}{System-level ($r$)} &  \multirow{1}{*}{8} & \underline{0.93} & \underline{0.93} & \underline{0.97} \\
 \multirow{1}{*}{Segment-level ($\tau$)} & 640 & 0.33 & 0.39 & 0.42\\
\bottomrule
\end{tabular}}
\caption{\label{tab:corr-style}
 Correlation of automatic metrics in style strength with human judgements. The underlined scores indicate $p$ $<$ 0.01.} 
\end{table}

\begin{figure}[t]
    \centering
    \includegraphics[scale=0.48]{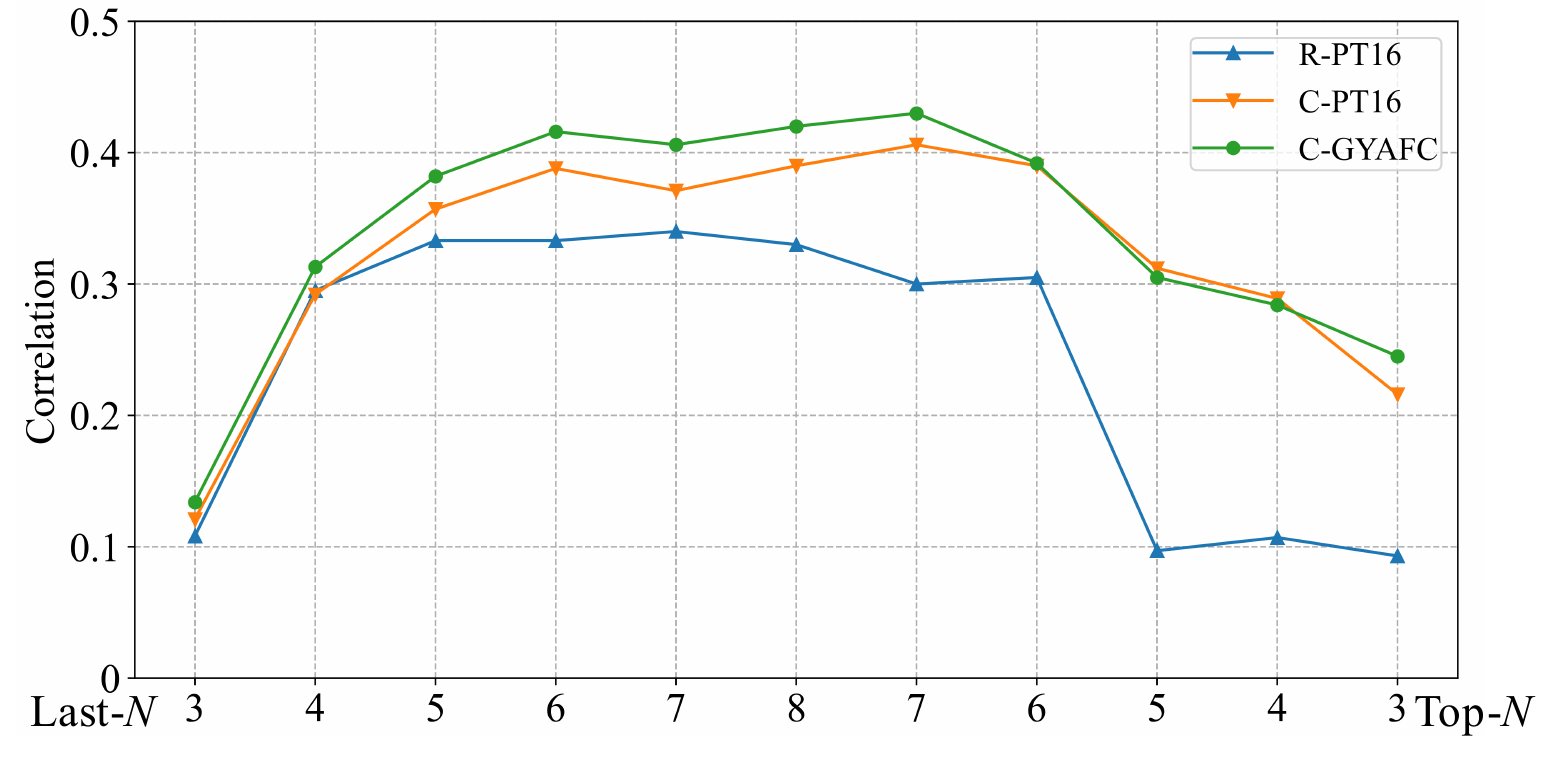}
    \caption{Kendall’s Tau-like correlation in style strength computed over the top-/last-$N$ systems which are sorted by human judgements.}
    \label{fig:corr-style}
\end{figure}

\subsection{Style Strength}

Table~\ref{tab:corr-style} shows the correlation of automatic metrics in style strength with human judgements. We see that C-GYAFC achieves the highest correlation at both system- and segment-level, R-PT16 and C-PT16 have the same system-level correlation score while the former has a slightly lower score at segment-level. Given that C-PT16 and C-GYAFC have close correlation scores while their performances on the test set are quite different, we also employ Pearson correlation to compute the segment-level result, and see rather different correlation scores (C-PT16 with 0.33 and C-GYAFC with 0.67). 
We think that evaluating the system outputs for a given source using C-PT16 and C-GYAFC results in similar scores ranking, so their Kendall’s Tau-like correlations are very close. 

In general, it is easier to evaluate systems which have large differences in quality, while it is more difficult when systems have similar quality. To assess the reliability of automatic metrics for close-quality systems, we first sort the systems based on human judgements, and plot the correlation of the top-/last-$N$ systems, with $N$ ranging from all systems to the best/worst three systems (Fig.~\ref{fig:corr-style}). 
We see that the correlation between automatic metrics and human judgements decreases as we decrease $N$ for both top-$N$ and last-$N$ systems, especially R-PT16 in the top-$N$ systems. Again we observe that C-GYAFC and C-PT16 have similar scores over the top-/last-$N$ systems. Overall, C-GYAFC appears to be the most stable model.

\begin{table*}[t]
\setlength{\tabcolsep}{4pt}
\resizebox{\linewidth}{!}{%
\centering
\footnotesize
\begin{tabular}{l|c|cccccccc|cccccccc}
\toprule[1.1pt]
  \multirow{2}{*}{\textbf{Systems}} & \multirow{2}{*}{\textbf{AVE. z}}  & \multicolumn{8}{c|}{\textbf{Source Sentence}} & \multicolumn{8}{c}{\textbf{Human Reference}}\\
  \cline{3-18}
  & & \rotatebox{90}{BLEU} & \rotatebox{90}{chrF} & \rotatebox{90}{ROUGE-L} & \rotatebox{90}{WMD} $\xdownarrow[0.65cm]$ & \rotatebox{90}{METEOR} & \rotatebox{90}{BERTScore} & \rotatebox{90}{BLEURT} & \rotatebox{90}{COMET-w} & \rotatebox{90}{BLEU} & \rotatebox{90}{chrF} & \rotatebox{90}{ROUGE-L} & \rotatebox{90}{WMD} $\xdownarrow[0.65cm]$ & \rotatebox{90}{METEOR} & \rotatebox{90}{BERTScore} & \rotatebox{90}{BLEURT} & \rotatebox{90}{COMET-w}\\
  \midrule
  Reference & 0.009 & 0.291 & 0.492 & 0.501 & 1.334 & 0.487 & 0.605 & 0.235 & 0.314  & - & - & - & - & - & - \\
  \hline
  HIGH    & \textbf{0.542} & 0.608 & 0.775 & 0.758 & 0.672 & 0.808 & \textbf{0.880} & \textbf{0.851} & 0.895 & 0.366 & 0.547 & 0.582 & 1.086 & 0.554 & 0.643 & 0.347 & 0.400\\
  NIU     & 0.491 & 0.637 & 0.772 & 0.769 & 0.652 & 0.808 & 0.873 & 0.818 & \textbf{0.899} & 0.376 & \textbf{0.560} & \textbf{0.605} & \textbf{1.036} & 0.567 & \textbf{0.649} & 0.373 & 0.418\\
  BART    & 0.370 & 0.514 & 0.688 & 0.692 & 0.840 & 0.724 & 0.798 & 0.687 & 0.752 & \textbf{0.382} & 0.555 & 0.596 & 1.053 & 0.573 & 0.646 & \textbf{0.388} & \textbf{0.425}\\
  IBT     & 0.337 & 0.543 & 0.711 & 0.717 & 0.782 & 0.749 & 0.838 & 0.744 & 0.813 & 0.373 & 0.550 & 0.582 & 1.094 & \textbf{0.574} & 0.635 & 0.350 & 0.391\\
  RAO     & 0.328 & \textbf{0.649} & \textbf{0.778} & \textbf{0.791} & \textbf{0.608} & \textbf{0.815} & 0.833 & 0.751 & 0.822 & 0.336 & 0.525 & 0.561 & 1.145 & 0.533 & 0.601 & 0.234 & 0.305\\
  ZHOU    &-0.659 & 0.610 & 0.717 & 0.765 & 0.758 & 0.770 & 0.739 & 0.189 & 0.318 & 0.253 & 0.461 & 0.494 & 1.351 & 0.469 & 0.508 &-0.200 &-0.125\\
  YI      &-0.669 & 0.547 & 0.684 & 0.731 & 0.823 & 0.728 & 0.716 & 0.148 & 0.320 & 0.288 & 0.483 & 0.517 & 1.307 & 0.491 & 0.524 &-0.154 &-0.059\\
  LUO     &-0.749 & 0.472 & 0.638 & 0.660 & 1.034 & 0.681 & 0.646 & 0.020 & 0.034 & 0.222 & 0.416 & 0.445 & 1.514 & 0.434 & 0.453 &-0.289 &-0.278\\
\bottomrule[1.1pt]
\end{tabular}}
\caption{\label{tab:results-metrics}
 Human evaluation (z-score) and automatic metrics in content preservation. Notes: (i) $\downarrow$ indicates the lower the score the better; (ii) COMET-w indicates that the input setting is not used.}
\end{table*}

\begin{figure}[t]
    \centering
    \includegraphics[scale=0.32]{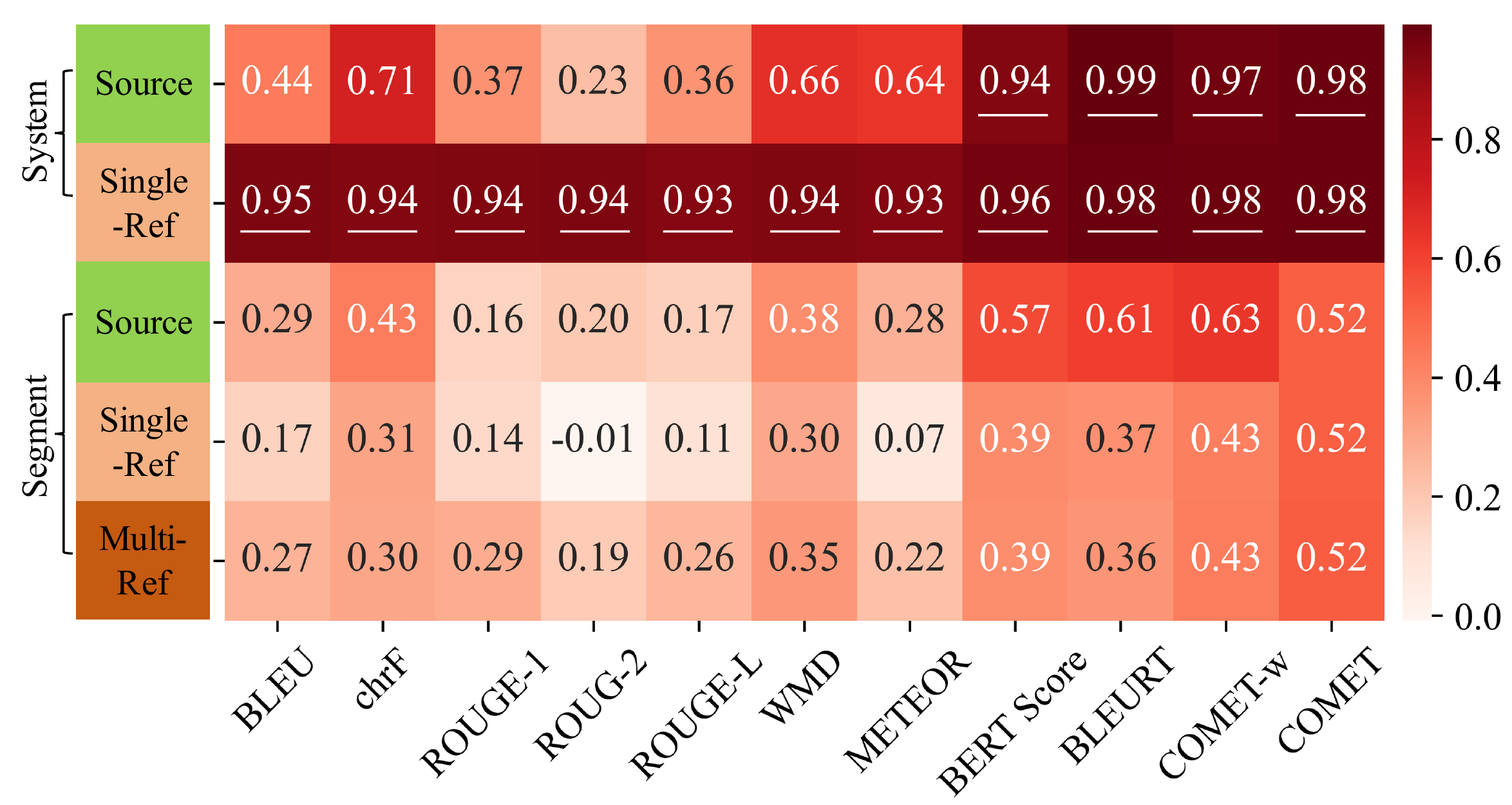}
    \caption{Correlations of automatic metrics  computed against source/reference in content preservation with human judgments. Underlining indicates $p<0.01$.}
    \label{fig:corr-content}
\end{figure}

\begin{table}[t]
\setlength{\tabcolsep}{4pt}
\resizebox{\linewidth}{!}{%
\centering
\footnotesize
\begin{tabular}{lcccccccccc}
\toprule[1.1pt]
   & \rotatebox{90}{BLEU} & \rotatebox{90}{chrF} & \rotatebox{90}{ROUGE-1} & \rotatebox{90}{ROUGE-2} & \rotatebox{90}{ROUGE-L} & \rotatebox{90}{WMD} & \rotatebox{90}{METEOR} & \rotatebox{90}{BERTScore} & \rotatebox{90}{BLEURT} & \rotatebox{90}{COMET-w}\\
 \hline
 Reference 2 & 0.28 & 0.37 & 0.33 & 0.10 & 0.36 & 0.46 & 0.21 & 0.59 & 0.61 & 0.61\\
 Reference 3 & 0.25 & 0.41 & 0.37 & 0.12 & 0.35 & 0.47 & 0.34 & 0.60 & 0.60 & 0.55\\
 Reference 4 & 0.37 & 0.41 & 0.46 & 0.24 & 0.46 & 0.49 & 0.31 & 0.60 & 0.56 & 0.62\\
\bottomrule[1.1pt]
\end{tabular}}
\caption{\label{tab:corr-reference}
Kendall’s Tau-like correlation between using the first human reference and other references for evaluation content preservation at segment-level.}
\end{table}

\begin{figure*}[t]
    \begin{minipage}[t]{0.52\linewidth}
    \subfigure[Automatic metrics results against source sentence.]{
      \includegraphics[scale=0.48]{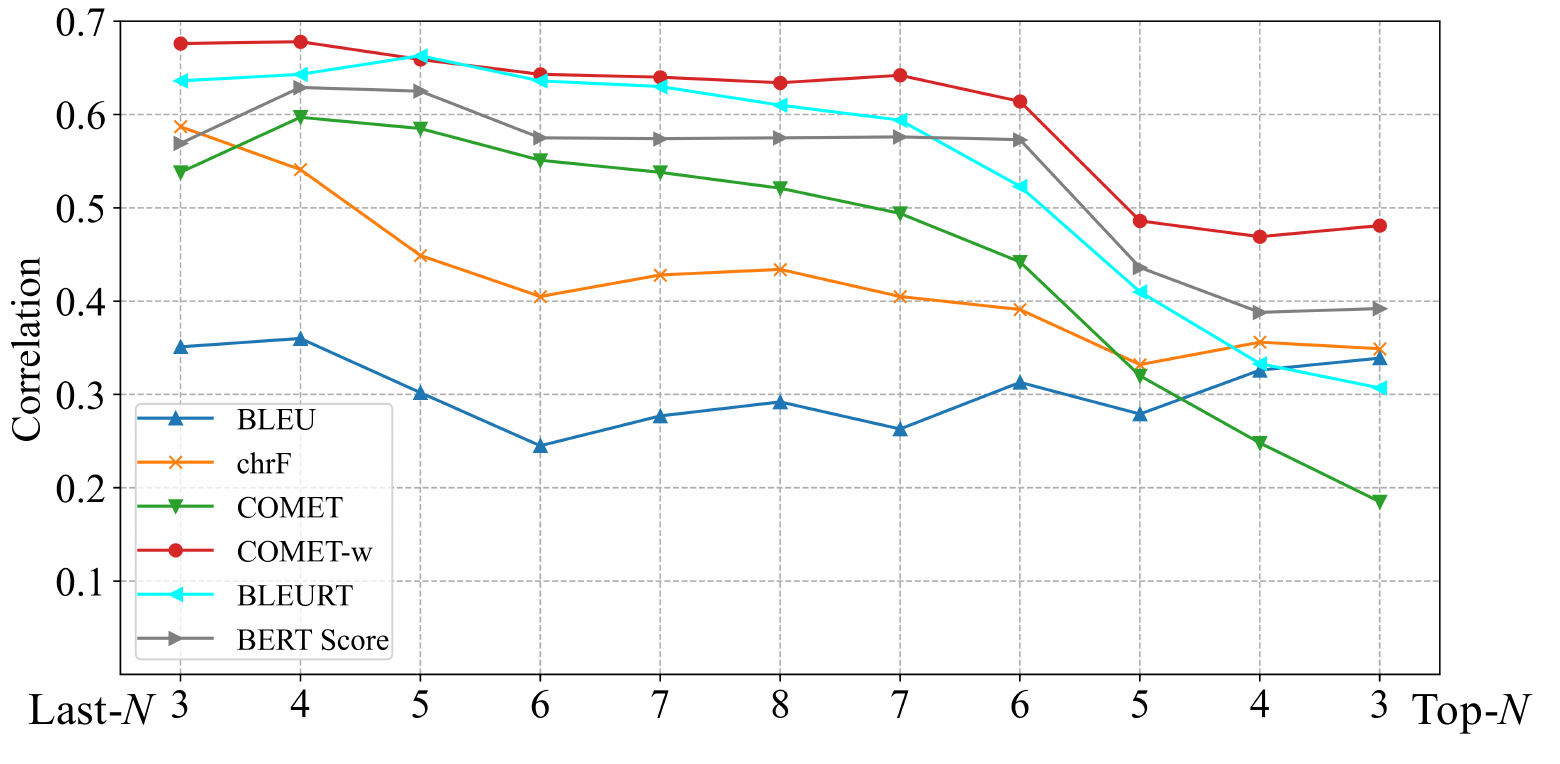}
      \label{fig:top-last-source}
    }
    \end{minipage}
    \begin{minipage}[t]{0.52\linewidth}
    \subfigure[Automatic metrics results against human reference.]{
        \includegraphics[scale=0.48]{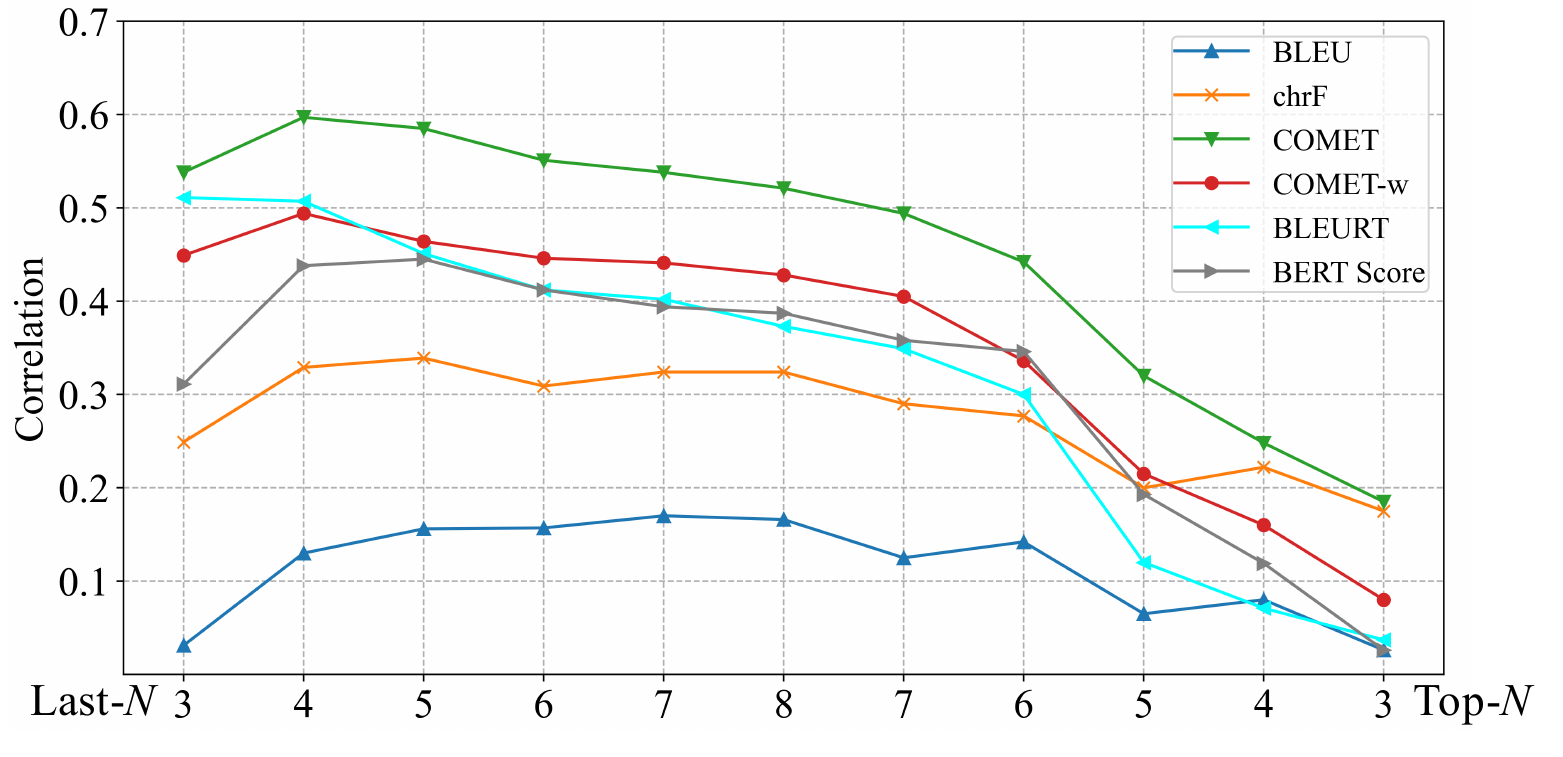} 
        \label{fig:top-last-reference}
    }
    \end{minipage}
    \caption{Kendall’s Tau-like correlation in content preservation computed over the top-/last-$N$ systems which are sorted by human judgements. }
    \label{fig:seg-corr}
\end{figure*}

\subsection{Content Preservation}

As mentioned in the Introduction, since a style-transformed output should not alter the meaning of the input, content preservation can be measured against the input itself, or against a human reference in the expected target style. However, metrics cannot be used interchangeably (Section~\ref{sub: metric}), as, for instance, the output is expected to have a higher $n$-gram overlap with the reference, while this is not desirable with respect to the input.

Table~\ref{tab:results-metrics} presents the results of human and automatic evaluation: all systems have a higher $n$-gram overlap (BLEU, chrF) with the source sentence than the human reference, indicating that existing models tend to copy from the input and lack diverse rewriting abilities. We also report the results for the reference against the source. Bearing in mind that the reference can be conceived as an optimal output, it is interesting to see that it does not score high in any metric, not even the learnable ones. This leaves some crucial open questions: how can these metrics be best used to assess content preservation in generated outputs? What are desirable scores?
We also observe that \texttt{RAO}'s system has the highest scores of surface-based metrics (e.g.\ BLEU) with the source sentence while its scores with learnable metrics (e.g.\ BLEURT) are lower than some other systems (e.g.\ \texttt{HIGH}). In the evaluation against the human reference, the system \texttt{BART} and \texttt{NIU} achieve better results on most metrics.

Figure~\ref{fig:corr-content} shows the correlations of content preservation metrics  with human judgments. For the system-level results, there is a big gap in correlation between source sentence and human reference for surface-based metrics (e.g.\ BLEU), but not for neural network based ones (e.g.\ COMET). Using the latter therefore seems to open up the possibility of automatically evaluating content without a human reference. It is interesting to see that the correlations of using source sentences at segment-level are all higher than using the human reference, and surface-based metrics of the latter correlate particularly poorly with human scores. We suggest two main reasons: (i) existing systems lack diverse rewriting ability given the source sentences, and the annotators rate the generated sentences comparing them to the source sentence, not to a reference
; (ii) human references are linguistically more diverse (e.g.\ word choice and order). The first one is not within the scope of this work. For the second aspect, we exploit the fact that we have multiple references available, and run the evaluation in a multi-reference setting; we observe that correlations for surface-based metrics improve as more variety is included, but not for neural ones. In Table~\ref{tab:corr-reference}, we see that learnable metrics using the first reference have higher correlation with other references than surface-based metrics.
Overall, learnable metrics always have the highest correlation scores in evaluating content preservation using source sentences or human references, while surface-based metrics generally require a multi-reference setting.

Similar to style strength, we plot the correlation of the top-/last-$N$ systems sorted by human judgements 
for the content aspect (Fig.~\ref{fig:seg-corr}). The correlation score between automatic metrics and human scores decreases as we decrease $N$ for the top-$N$ systems while this shows stability for the last-$N$ systems. This suggests that evaluating high-quality TST systems is more challenging than evaluating low-quality systems. Again, we see that the correlation when using the source sentence has better stability than when using human references. Although BLEU and charF show stable performances, their correlations are lower than those by other metrics in most cases. Regardless of whether we use human references or source sentences, COMET(-w) generally has the highest correlation scores with human judgements under different conditions.

\subsection{Fluency}

\begin{table}[t]
\setlength{\tabcolsep}{4pt}
\resizebox{\linewidth}{!}{%
\centering
\footnotesize
\begin{tabular}{lccc}
\toprule[1.1pt]
  & \textbf{N} & \textbf{Informal-to-Formal} &  \textbf{Formal-to-Informal} \\
 \hline
  \multirow{1}{*}{System-level ($r$)} &  \multirow{1}{*}{8} & \underline{0.96} & 0.65 \\
 \multirow{1}{*}{Segment-level ($\tau$)} & 320 & 0.52 & 0.35\\
\bottomrule[1.1pt]
\end{tabular}}
\caption{\label{tab:corr-fluency}
 Absolute correlation of automatic metrics in fluency with human judgements. The underlined scores indicate $p$ $<$ 0.01.}
\end{table}

\begin{table}[t]
\setlength{\tabcolsep}{4pt}
\resizebox{\linewidth}{!}{%
\centering
\footnotesize
\begin{tabular}{l|ccc|ccc}
\toprule[1.1pt]
  & \multicolumn{3}{c|}{\textbf{Informal-to-Formal}} & \multicolumn{3}{c}{\textbf{Formal-to-Informal}} \\
  \cline{2-7}
  & GPT2-Inf & GPT2-For & $r$ & GPT2-Inf & GPT2-For & $r$ \\
  \midrule
  Source    & 76 & 143 & -    & 87  & 68  & -\\
  Reference & 60 & 37  & 0.21 & 115 & 270 & 0.13\\
  \hline
  BART & 34 & \textbf{26} & 0.33 & \textbf{24} & \textbf{28} & 0.02\\
  IBT & \textbf{32} & \textbf{26} & 0.32 & 33 & 40 & 0.17\\
  NIU & 43 & 37 & 0.30 & 71 & 75 & 0.03\\
  HIGH & 41 & 35 & \textbf{0.62} & 80 & 75 & 0.00\\
  RAO & 54 & 57 & 0.33 & 54 & 55 & 0.02\\
  ZHOU & 189 & 218 & 0.36 & 103 & 111 & \textbf{0.42}\\
  YI & 160 & 182 & 0.31 & 205 & 436 & 0.27\\
  LUO & 128 & 152 & 0.43 & 6962 & 8191 & 0.17\\
\bottomrule[1.1pt]
\end{tabular}}
\caption{\label{tab:ppl-fluency}
 Results of GPT-2 based perplexity scores and their absolute Pearson correlation with human judgements at segment-level. Notes: (i) GPT2-Inf and GPT2-For are fine-tuned with informal sentences and formal sentences, respectively; (ii) the correlation is calculated using the perplexity of GPT-2 in the target style with human judgment.}
\end{table}

Table~\ref{tab:corr-fluency} shows the absolute correlation of fluency metrics with human judgements. Unsurprisingly, we see that GPT-2 based perplexity correlates better with human scores in the direction informal-to-formal than in the opposite one, at both system- and segment-level. In general, a ``good'' formal sentence should be fluent, while an informal sentence might as well not be, and there can be varied perceptions by people. Indeed, we see higher IAA scores in the informal-to-formal direction (informal-to-formal: 0.70 vs informal-to-formal: 0.63). 
Table 6 presents the results of correlations and perplexity scores of GPT-2 in the two transfer directions for each system. The perplexity scores for most sentences are in the \textit{correct} place, i.e. the scores from GPT2-Inf are higher than those from GPT2-For for the informal sentences, and viceversa. However, we also observe that the correlations of informal-to-formal for each system (except \texttt{ZHOU}) are higher than those for the formal-to-informal direction. This confirms our hypothesis that assessing the fluency of informal sentences is not that obvious 
even for humans.

\subsection{Broader Implications for Style Transfer}
\label{sub:task-nature}

\begin{figure}[t]
    \centering
    \includegraphics[scale=0.43]{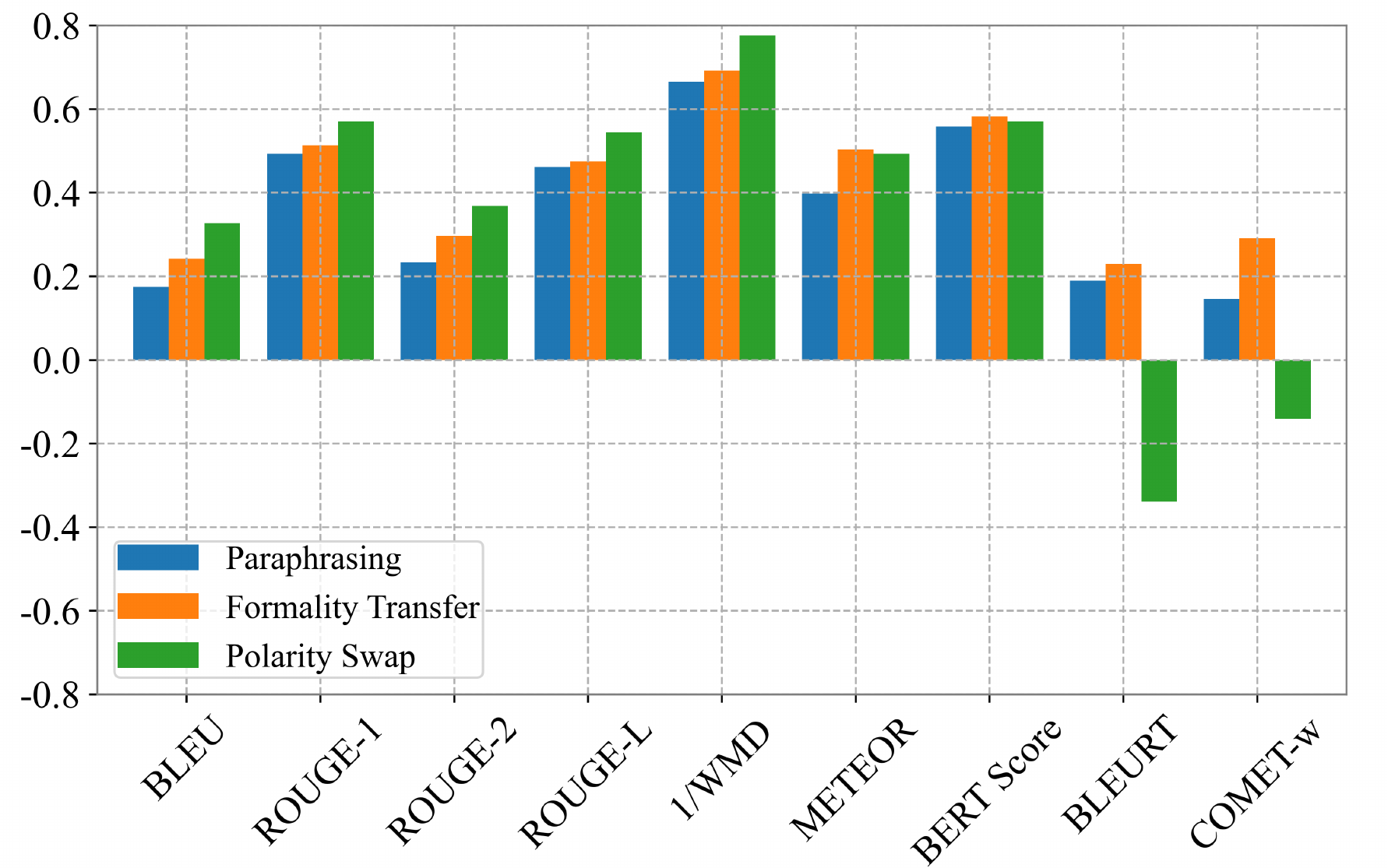}
    \caption{The distance between the source and target sentences as measured by content-related metrics.}
    \label{fig:distance}
\end{figure}

We have focused here on formality transfer, but polarity swap is also commonly defined as a style transfer task. In previous work, we have  suggested that these tasks are intrinsically different, especially in terms of content preservation, since while formality transfer is somewhat akin to paraphrasing, in polarity swap the meaning is substantially altered \citep{lai-etal-2021-generic}. This would imply that content-measuring metrics could not be used in the same way in the two tasks. 

We further peek here into this issue, in view of future work that should evaluate metrics for the assessment of polarity swap, too,  and show in 
Figure~\ref{fig:distance} the use of different  metrics to measure the distance between the source and target sentences for paraphrasing, formality transfer, and polarity swap. Using $n$-gram based metrics, we see that the distance between source and target sentences in polarity swap is closer than in the other two tasks. With learnable metrics, instead, we see that source and target sentences for polarity swap are quite distant. Formality transfer shows overall the same trend as paraphrasing in all metrics, suggesting that it's much more of a content-preserving paraphrase-like task than polarity swap, and metrics should be selected accordingly. Future work will explore how to best use them in polarity swap under different settings (using source vs reference, for example).

\section{Conclusion}
We have considered a wide range of automatic metrics on the three evaluation aspects of formality transfer, and assessed them against human judgements that we have elicited.

For \textbf{style strength}, we have compared the style classifiers and regressor in the setting of using the same raw data for training (with a binary label for classification and continuous scores for regression), as well as classifiers with different performances. We have observed that there is little difference among them when evaluating multiple TST systems. However, the style regressor performs worse when evaluating high-quality TST systems. For classifiers with different performances, we recommend the one with the highest performance since it 
results in the highest overall Pearson correlation with human judgements.


To assess \textbf{content preservation}, we have explored different kinds of automatic metrics using the source or reference(s), and have observed the follwoing: (i) 
if using the
source sentence, we strongly recommend employing learnable metrics since 
their correlation in that condition is much higher than those of traditional surface-based metrics (which are not indicative, since high scores correspond to not changing the input, hence not performing the task); still, the question of how scores should be interpreted and what score ranges are desirable remains open;
(ii) most metrics are reliable to be used to measure and compare the performances at system-level when a human reference is available; (iii) however, we do not recommend to use surface-base metrics to measure sentence-level comparisons, especially with only one reference. Overall, learnable metrics seem to provide a more reliable measurement. 

For \textbf{fluency}, perplexity can be used for evaluating the informal-to-formal direction, either at system- or segment-level, while it is clearly less reliable for the opposite direction, and it remains to be investigated how to best perform evaluation in this transfer direction, considering the wide variability of acceptable outputs.

This study focuses on formality transfer, and offers a better understanding of automatic evaluation thanks to the comprehensive correlations with human judgments herein conducted. 
However, the findings may not generalise to other tasks usually considered similar, such as polarity swap. To this end, future dedicated work will be required.

\section*{Acknowledgments}

This work was partly funded by the China Scholarship Council (CSC). We are very grateful to the anonymous reviewers for their useful comments, especially in connection to closely related work, which contributed to strengthening this paper. We would also like to thank the Center for Information Technology of the University of Groningen for their support and for providing access to the Peregrine high performance computing cluster. We thank the annotators as well as Ana Guerberof Arenas and Amy Isard for testing, and helping us to improve, a preliminary version of the survey.


\bibliography{anthology,custom}
\bibliographystyle{acl_natbib}
\clearpage
\appendix
\section{\Large Appendices: \\~ \\ }
\label{sec:appendix}

\setcounter{table}{0}
\renewcommand{\thetable}{A.\arabic{table}}
\setcounter{figure}{0}
\renewcommand{\thefigure}{A.\arabic{figure}}

These Appendices include: (i) evaluated systems (\ref{app:formality-system}); (ii) implementation details for automatic metrics (\ref{app:details-metrics}); and (iii) annotation guidelines and interface (\ref{app:interface})\vspace*{0.5cm}.

\subsection{Evaluated Systems}
\label{app:formality-system}

Table~\ref{tab:formality-system-ranking} presents the systems' ranking based on the human judgements. We use eight published systems of different sorts (including state-of-the-art models). For \textbf{supervised approaches}, we include the following systems:
\begin{itemize}[leftmargin=*]
\itemsep 0in
\item RAO~\citep{rao-tetreault-2018}: A copy-enriched NMT model trained on the rule-processed data and the additional forward and backward translations produced by the PBMT model;

\item NIU~\citep{niu-etal-2018-multi}: A bi-directional model trained on formality-tagged bilingual data using multi-task learning;

\item BART~\citep{lai-etal-2021}: Fine-tuning a pre-trained model BART with gold parallel data and reward strategies; 

\item HIGH~\citep{lai-etal-2021-generic}: Fine-tuning BART with high-quality synthetic parallel data and reward strategies.
\end{itemize}

\noindent For \textbf{unsupervised approaches}, we include the following systems:

\begin{itemize}[leftmargin=*]
\itemsep 0in
\item LUO~\citep{fuli-2019}: A dual reinforcement learning framework that directly transforms the style of the text via a one-step mapping model without parallel data; 

\item YI~\citep{xiaoyuan-ijcai}: A style instance supported method that learns a more discriminative and expressive latent space to enhance style signals and make a better balance between style and content; 

\item Zhou~\citep{zhou-etal-2020-exploring}: An attentional seq2seq model that pre-trains the model to reconstruct the source sentence and re-predict its word-level style relevance;

\item IBT~\citep{lai-etal-2021-generic}: An iterative back-translation framework based on the pre-trained seq2seq model BART.
\end{itemize}

\noindent Table~\ref{tab:results-content} presents automatic evaluation results in content preservation. 

\subsection{Implementation Details for Automatic Metrics}
\label{app:details-metrics}

\begin{itemize}[leftmargin=*]
\itemsep 0in

\item BLEU: We adopt \texttt{sentence\_bleu} of the NLTK library with a smoothing function to compute the segment-level score, and \texttt{multi-bleu.perl} with default settings for system-level.\footnote{\url{https://www.nltk.org/}}

\item chrF: Following~\citet{briakou-etal-2021-evaluating}, we use use \texttt{sentence\_chrf} of the open-sourced implementation sacreBLEU.\footnote{\url{https://github.com/mjpost/sacrebleu}}

\item ROUGE: We use the open-sourced implementations Rouge.\footnote{\url{https://github.com/pltrdy/rouge}}

\item WMD: We employ the \texttt{gensim} library and word embedding \texttt{googlenews-vectors-} \texttt{negative300.bin}.\footnote{\url{https://radimrehurek.com/gensim/index.html}}

\item METEOR: We adopt the NLTK library.

\item BERTScore: We use the official implementation with a rescaling function.\footnote{\url{https://github.com/Tiiiger/bert_score}} 

\item BLEURT:  We use the official checkpoint of \texttt{bleurt-large-512}.\footnote{\url{https://github.com/google-research/bleurt}}

\item COMET: We adopt the official checkpoint of \texttt{wmt-large-da-estimator-1719}.\footnote{\url{https://github.com/Unbabel/COMET}} COMET-QE is a referenceless metric that uses source and output only. But we found that it yielded lower correlations with human judgements than COMET in our evaluations. This may be because the input and output are different languages in COMET-QE training.

\item Style and Fluency: All experiments are implemented atop Transformers~\citep{wolf-etal-2020-transformers} using BERT base model (cased) for style and GPT-2 base model for fluency. We fine-tune models using the Adam optimiser~\citep{kingma2017adam} with learning rate of 1e-5 for BERT and 3e-5 for GPT-2, with a batch size of 32 for all experiments.

\end{itemize}

\subsection{Annotation Guidelines and Interface}
\label{app:interface}
\begin{figure*}[!t]
    \begin{minipage}[t]{0.5\linewidth}
    \centering
    \subfigure[A screenshot of task guidelines.]{
      \includegraphics[scale=0.57]{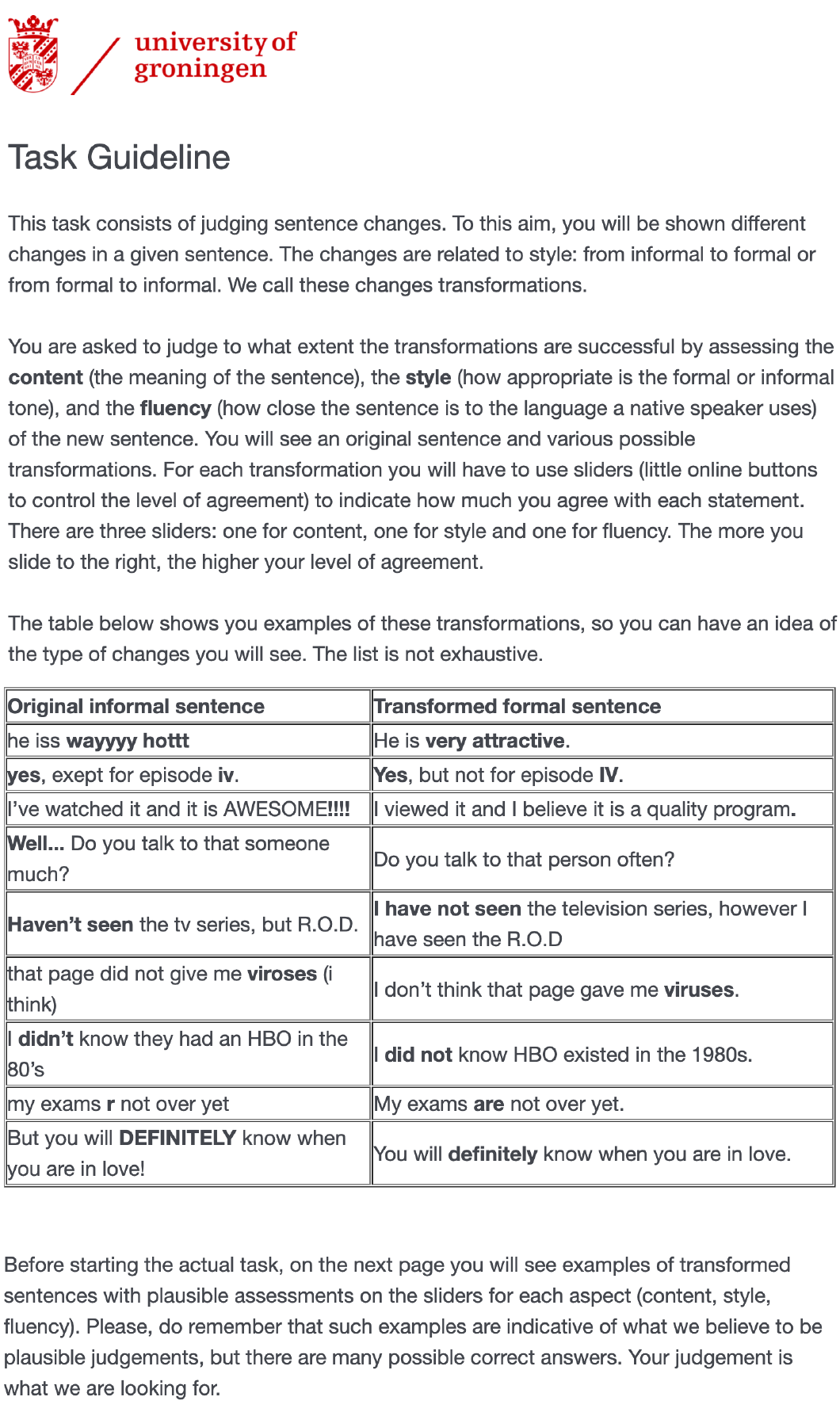}
      \label{fig:guideline}
    }
    \end{minipage}
    \begin{minipage}[t]{0.5\linewidth}
    \centering
    \subfigure[A screenshot of annotation interface.]{
        \includegraphics[scale=0.65]{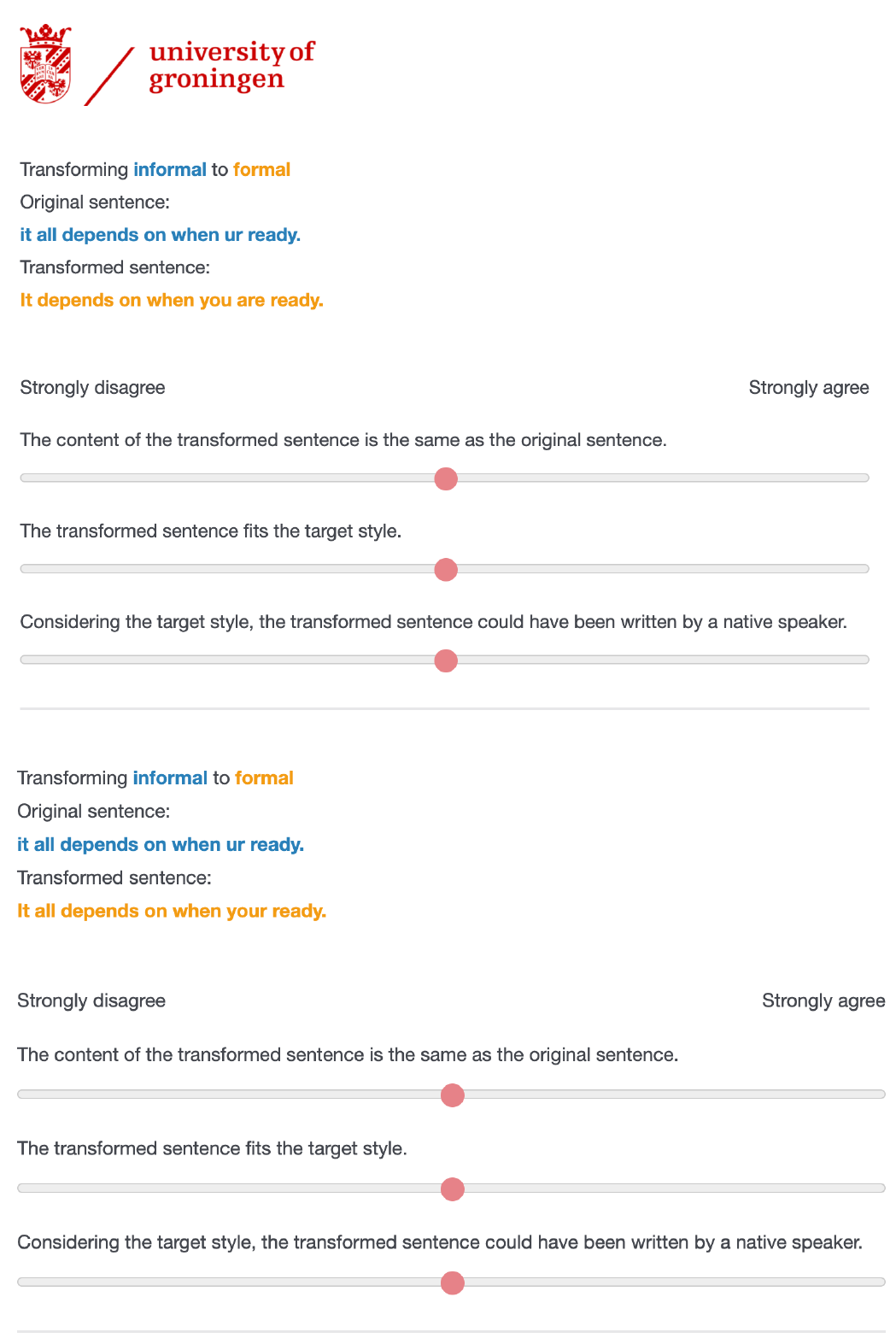} 
        \label{fig:annotation}
    }
    \end{minipage}
    \caption{Screenshots of our interface.}
    \label{fig:interface}
\end{figure*}

Figure~\ref{fig:interface} show  the screenshots of task guidelines and annotation interface.

\begin{table*}[!t]
\setlength{\tabcolsep}{4pt}
\centering
\footnotesize
\begin{tabular}{lccc|lccc|lccc}
\toprule
\multicolumn{4}{c|}{\textbf{Style}} & \multicolumn{4}{c|}{\textbf{Content}} & \multicolumn{4}{c}{\textbf{Fluency}} \\
\hline
 \textbf{System} & \textbf{Rank} & \textbf{AVE. s} & \textbf{AVE. z} & \textbf{System} & \textbf{Rank} & \textbf{AVE. s} & \textbf{AVE. z} & \textbf{System} & \textbf{Rank} & \textbf{AVE. s} & \textbf{AVE. z}\\
 \midrule
  BART & 1 & 82.7 & 0.494  & HIGH & 1 & 92.4 & 0.542  & BART & 1 & 87.8 & 0.540\\
  REF  & 2 & 82.3 & 0.469  & NIU  & 2 & 90.7 & 0.491  & IBT  & 2 & 86.0 & 0.491\\
  IBT  & 3 & 80.1 & 0.407  & BART & 3 & 86.5 & 0.370  & NIU  & 3 & 84.9 & 0.463\\
  NIU  & 4 & 76.9 & 0.297  & IBT  & 4 & 85.1 & 0.337  & HIGH & 4 & 83.3 & 0.420\\
  HIGH & 5 & 76.3 & 0.293  & RAO  & 5 & 84.7 & 0.328  & REF  & 5 & 82.4 & 0.385\\
  RAO  & 6 & 70.2 & 0.085  & REF  & 6 & 73.6 & 0.009  & RAO  & 6 & 77.3 & 0.247\\
  YI   & 7 & 51.1 & -0.588 & ZHOU & 7 & 50.9 & -0.659 & ZHOU & 7 & 45.1 & -0.717\\
  ZHOU & 8 & 47.2 & -0.726 & YI   & 8 & 50.5 & -0.669 & YI   & 8 & 38.6 & -0.903\\
  LUO  & 9 & 46.7 & -0.731 & LUO  & 9 & 47.6 & -0.749 & LUO  & 9 & 37.9 & -0.926\\
\bottomrule
\end{tabular}
\caption{\label{tab:formality-system-ranking}
Results based on original human evaluation and z-score.}
\end{table*}

\begin{table*}[!t]
\resizebox{\linewidth}{!}{%
\centering
\footnotesize
\begin{tabular}{l|cccccccccc|cccccccccc}
\toprule[1.1pt]
  & \rotatebox{90}{BLEU} & \rotatebox{90}{chrF} & \rotatebox{90}{ROUGE-1} & \rotatebox{90}{ROUGE-2} & \rotatebox{90}{ROUGE-L} & \rotatebox{90}{WMD} $\xdownarrow[0.65cm]$ & \rotatebox{90}{METEOR} & \rotatebox{90}{BERTScore} & \rotatebox{90}{BLEURT} & \rotatebox{90}{COMET-w} & \rotatebox{90}{BLEU} & \rotatebox{90}{chrF} & \rotatebox{90}{ROUGE-1} & \rotatebox{90}{ROUGE-2} & \rotatebox{90}{ROUGE-L} & \rotatebox{90}{WMD} $\xdownarrow[0.65cm]$ & \rotatebox{90}{METEOR} & \rotatebox{90}{BERTScore} & \rotatebox{90}{BLEURT} & \rotatebox{90}{COMET-w}\\
  \toprule[0.6pt]
  \textbf{Systems} & \multicolumn{10}{c|}{\textbf{Reference 1}} & \multicolumn{10}{c}{\textbf{Reference 2}}\\
  \hline
  Reference & 0.291 & 0.492 & 0.533 & 0.307 & 0.501 & 1.334 & 0.487 & 0.605 & 0.235 & 0.314 & 0.231 & 0.459 & 0.494 & 0.259 & 0.449 & 1.469 & 0.444 & 0.565 & 0.155 & 0.202\\
  \hline
  HIGH    & 0.366 & 0.547 & 0.624 & 0.401 & 0.582 & 1.086 & 0.554 & 0.643 & 0.347 & 0.400 & 0.300 & 0.515 & 0.564 & 0.342 & 0.512 & 1.260 & 0.521 & 0.605 & 0.317 & 0.289\\
  NIU     & 0.376 & 0.560 & 0.646 & 0.434 & 0.605 & 1.036 & 0.567 & 0.649 & 0.373 & 0.418 & 0.333 & 0.525 & 0.578 & 0.369 & 0.526 & 1.202 & 0.538 & 0.617 & 0.329 & 0.286\\
  BART    & 0.382 & 0.555 & 0.632 & 0.412 & 0.596 & 1.053 & 0.573 & 0.646 & 0.388 & 0.425 & 0.305 & 0.511 & 0.561 & 0.349 & 0.513 & 1.278 & 0.526 & 0.605 & 0.353 & 0.279\\
  IBT     & 0.373 & 0.550 & 0.620 & 0.404 & 0.582 & 1.094 & 0.574 & 0.635 & 0.350 & 0.391 & 0.291 & 0.503 & 0.553 & 0.335 & 0.503 & 1.289 & 0.512 & 0.595 & 0.305 & 0.271\\
  RAO     & 0.336 & 0.525 & 0.602 & 0.367 & 0.561 & 1.145 & 0.533 & 0.601 & 0.234 & 0.305 & 0.297 & 0.505 & 0.556 & 0.344 & 0.512 & 1.281 & 0.512 & 0.568 & 0.200 & 0.196\\
  ZHOU    & 0.253 & 0.461 & 0.536 & 0.300 & 0.494 & 1.351 & 0.469 & 0.508 &-0.200 &-0.125 & 0.245 & 0.451 & 0.495 & 0.271 & 0.444 & 1.488 & 0.476 & 0.478 & -0.206 & -0.212\\
  YI      & 0.288 & 0.483 & 0.551 & 0.324 & 0.517 & 1.307 & 0.491 & 0.524 &-0.154 &-0.059 & 0.225 & 0.443 & 0.497 & 0.263 & 0.454 & 1.475 & 0.457 & 0.488 & -0.203 & -0.167\\
  LUO     & 0.222 & 0.416 & 0.483 & 0.272 & 0.445 & 1.514 & 0.434 & 0.453 &-0.289 &-0.278 & 0.189 & 0.381 & 0.419 & 0.209 & 0.378 & 1.694 & 0.389 & 0.425 & -0.266 & -0.368 \\
  \toprule[0.6pt]
  \textbf{Systems} & \multicolumn{10}{c|}{\textbf{Reference 3}} & \multicolumn{10}{c}{\textbf{Reference 4}}\\
  \hline
  Reference & 0.213 & 0.442 & 0.472 & 0.231 & 0.434 & 1.537 & 0.433 & 0.567 & 0.102 & 0.190 & 0.231 & 0.459 & 0.505 & 0.261 & 0.461 & 1.438 & 0.466 & 0.595 & 0.224 & 0.293\\
  \hline
  HIGH    & 0.316 & 0.513 & 0.566 & 0.340 & 0.528 & 1.229 & 0.506 & 0.617 & 0.236 & 0.326 & 0.295 & 0.511 & 0.585 & 0.343 & 0.535 & 1.227 & 0.526 & 0.634 & 0.327 & 0.412\\
  NIU     & 0.325 & 0.509 & 0.574 & 0.351 & 0.534 & 1.232 & 0.505 & 0.612 & 0.257 & 0.309 & 0.310 & 0.518 & 0.607 & 0.365 & 0.552 & 1.173 & 0.548 & 0.637 & 0.349 & 0.413\\
  BART    & 0.341 & 0.517 & 0.577 & 0.361 & 0.539 & 1.208 & 0.526 & 0.617 & 0.274 & 0.354 & 0.327 & 0.532 & 0.621 & 0.384 & 0.574 & 1.128 & 0.565 & 0.655 & 0.405 & 0.447\\
  IBT     & 0.307 & 0.514 & 0.570 & 0.344 & 0.531 & 1.220 & 0.522 & 0.614 & 0.267 & 0.328 & 0.316 & 0.520 & 0.592 & 0.363 & 0.543 & 1.217 & 0.534 & 0.632 & 0.332 & 0.388\\
  RAO     & 0.293 & 0.499 & 0.556 & 0.329 & 0.511 & 1.288 & 0.493 & 0.574 & 0.140 & 0.252 & 0.293 & 0.505 & 0.577 & 0.336 & 0.526 & 1.234 & 0.541 & 0.600 & 0.250 & 0.315\\
  ZHOU    & 0.227 & 0.419 & 0.478 & 0.245 & 0.438 & 1.496 & 0.421 & 0.489 &-0.257 &-0.186 & 0.210 & 0.425 & 0.507 & 0.248 & 0.453 & 1.451 & 0.448 & 0.507 &-0.212 &-0.162\\
  YI      & 0.220 & 0.436 & 0.487 & 0.255 & 0.449 & 1.477 & 0.416 & 0.488 &-0.263 &-0.149 & 0.204 & 0.432 & 0.501 & 0.250 & 0.458 & 1.466 & 0.430 & 0.509 &-0.182 &-0.086\\
  LUO     & 0.189 & 0.380 & 0.422 & 0.244 & 0.390 & 1.671 & 0.371 & 0.431 &-0.346 &-0.356 & 0.197 & 0.393 & 0.458 & 0.243 & 0.410 & 1.591 & 0.420 & 0.451 &-0.282 &-0.317\\
\bottomrule[1.1pt]
\end{tabular}}
\caption{\label{tab:results-content}
 Automatic evaluation results in content preservation. Notes: (i) the results of Reference is the distance between source and reference sentence measuring by metrics; (ii) $\downarrow$ indicates the lower score is better.}
\end{table*}

\end{document}